\documentclass[12pt,a4paper]{article}
\usepackage{amsmath}
\usepackage{graphicx}
\usepackage[colorlinks=true,
    linkcolor=black,
    citecolor=black,
    urlcolor=black]{hyperref}
\usepackage{cite}
\usepackage{geometry}
\usepackage{tikz}
\usetikzlibrary{calc}
\usepackage{chngcntr}
\counterwithin{figure}{section}
\begin{document}
\begin{titlepage}
\thispagestyle{empty} % delete page number
\begin{tikzpicture}[remember picture,overlay]
    \node at (current page.center) {
        \includegraphics[width=\paperwidth,height=\paperheight]{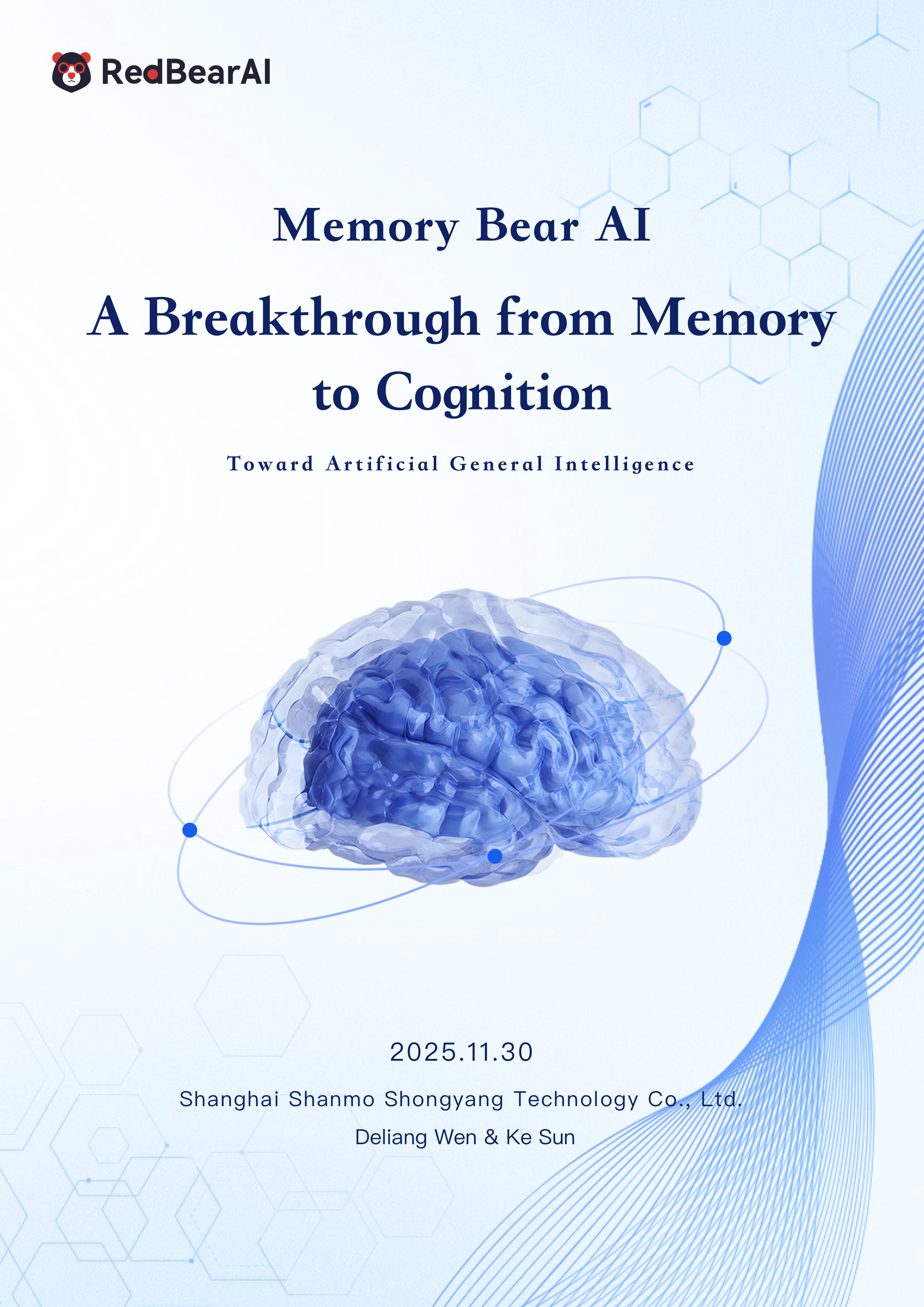}
    };
\end{tikzpicture}
\end{titlepage}

%\maketitle

\begin{abstract}
Large language models (LLMs) exhibit inherent memory limitations, including restricted context windows~\cite{ref1}, long-term knowledge forgetting, redundant information accumulation, and hallucination generation~\cite{ref2}, These limitations pose significant challenges for sustained dialogue and personalized services~\cite{ref3}. 

This paper introduces Memory Bear, a system that constructs a human-like memory architecture grounded in principles from cognitive science. By integrating multimodal information perception, dynamic memory maintenance, and adaptive cognitive services, Memory Bear achieves a full-chain reconstruction of LLM memory mechanisms for large language models.

Evaluated across domains including healthcare, enterprise operations, and education, Memory Bear demonstrates substantial engineering innovation and consistent performance gains. The system significantly improves knowledge fidelity and retrieval efficiency in long-term conversations, reduces hallucination rates, and enhances contextual adaptability and reasoning capability through tight integration of memory and cognition. 

Experimental results show that, compared with existing approaches (e.g., Mem0, MemGPT, Graphiti), Memory Bear outperforms them across key metrics—including accuracy, token efficiency, and response latency—marking a crucial step forward in advancing AI from ``memory'' to ``cognition.''
\end{abstract}
\begin{figure}[htbp]
    \centering
    \includegraphics[width=0.9\textwidth]{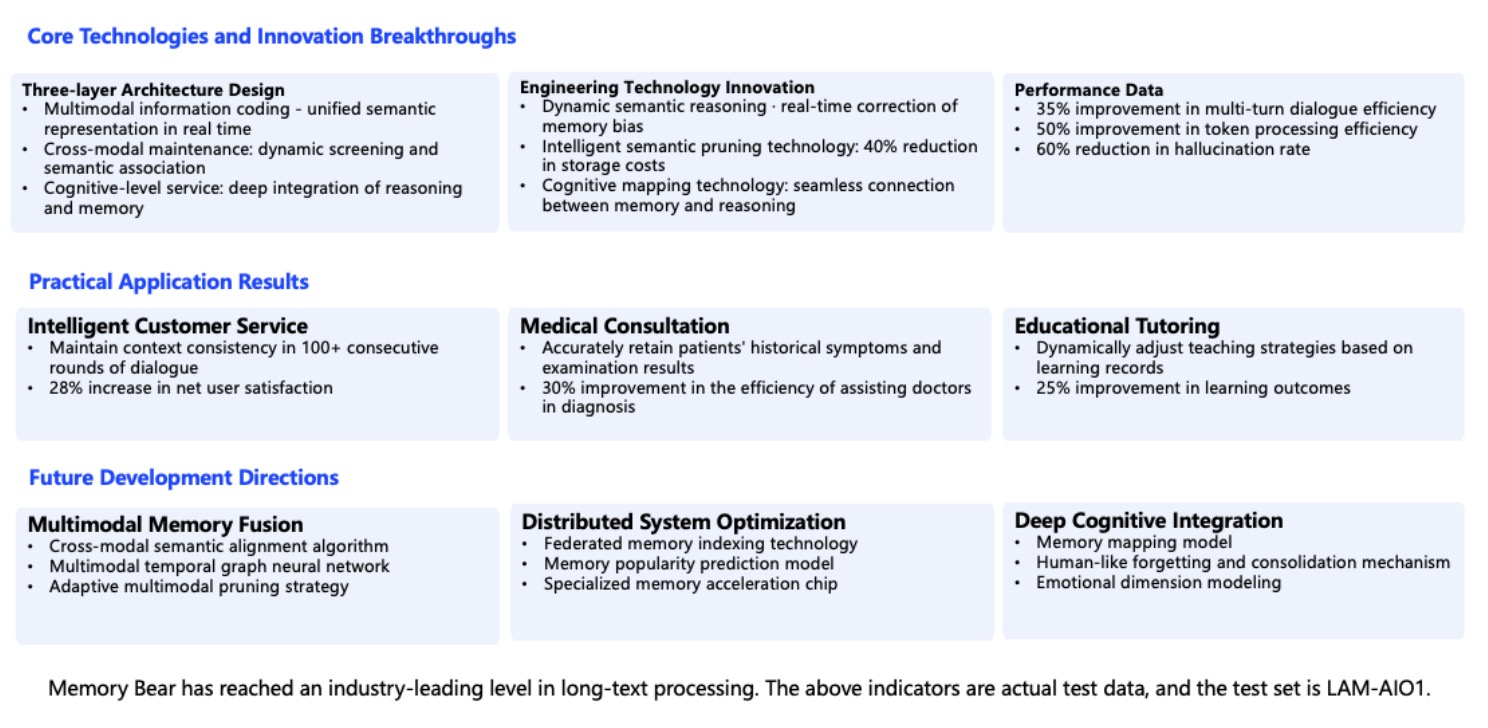}
    \caption{Core technologies, application results, and future development directions of Memory Bear.}
    \label{fig:1-1}
\end{figure}

%\begin{figure}[ht]
%    \centering
%    \includegraphics[width=0.95\textwidth]{Figure_1-1.jpg}
%   \caption{Core technologies, application results, and future development directions of Memory Bear.}
%    \label{fig:1-1}
%\end{figure}

\newpage
\tableofcontents

\newpage
\section{Problem Background}

Large language models (LLMs) face persistent challenges in knowledge memory and retrieval,
which have emerged as a "last mile” bottleneck in their progression toward higher intelligence\cite{ref4}. These challenges can be broadly characterized by context window limitations, long-term forgetting, context drift, token redundancy and memory hallucinations.

Context window limitations fundamentally limit LLM's ability to sustain long-term dialogues.
When interactions exceed the maximum window length, early but critical information is displaced by 
subsequent content, leading to the phenomenon of "forgetting what was said earlier\cite{ref5}.”
Although extending context length can temporarily mitigate this issue, it merely delays information loss rather than addressing the underlying memory deficiency\cite{ref6}. Long-term forgetting persists even in extended-context models, as information that falls outside the active window becomes inaccessible. Consequently, knowledge accumulated earlier in the interaction cannot be reliably retained or retrieved.

Context drift accumulates during long dialogue generation. As interactions progress, models may 
gradually deviate from the original topic or compound earlier erroneous assumptions\cite{ref7}. Token redundancy leads to inefficiency
and high costs. Traditional approaches often rely on repeatedly concatenating large volumes of historical
information into prompts, inflating token usage and distracting model attention by irrelevant information\cite{ref8}. When lacking effective memory mechanisms, LLMs must process thousands of context tokens without reliably extracting salient knowledge, further increasing response latency and computational overhead\cite{ref9}.

Memory hallucinations remain a critical concern\cite{ref10}. During storage and retrieval, existing AI memory modules frequently exhibit information fabrication, errors, conflicts, or omissions.
When relevent information is unavailable in the immediate context, models may fabricate plausible-sounding but incorrect responses. These problems are amplified in multi-agent collaboration scenarios—each agent operates independently, forming "memory silos” that force users to repeatedly provide 
the same information. In summary, limited long-term memory has become a key bottleneck 
restricting LLMs’ continuous, coherent, and personalized interactions.

Meanwhile, the demand for long-term memory systems in cognitive-level AI services is becoming increasingly urgent. In high-stakes domains such as healthcare, enterprise management, and education, real-world AI applications have evolved beyond short-term context processing and now require human-like memory–cognition integration. For example, in healthcare, chronic disease management often spans months or years, requiring AI systems to continuously track patients’ medication tolerance and disease progression\cite{ref11}. In enterprise settings, intelligent assistants must retrieve project records, business decision histories, and cross-team collaboration information from 6–10 months prior. Similarly, in educational environments\cite{ref12}, AI tutors are expected to reference students’ long-term learning trajectories, knowledge gaps, and evolving interests over periods exceeding 8 months in order to deliver genuinely personalized instruction\cite{ref13}.

These cross-cycle tasks highly depend on stable and consistent long-term memory mechanisms. 
However, existing memory approaches commonly suffer from long-term memory loss and context 
continuity breaks, making it difficult for AI systems to maintain task coherence across extend interactions. As a result, practitioners often resort to extensive manual prompt engineering to preserve context consistency, leading to high adaptation costs and unstable system behavior.

In response to these limitations, both academia and industry have increasingly recognized the importance of enhancing AI memory architectures. Human cognition naturally leverages long-term memory to sustain coherent conversations and dynamically update knowledge structures. In contrast, current LLM systems lack of persistent memory mechanisms, often forgetting user preferences, repeating questions, or generating contradictory responses--thereby undermining user experience and trust. Consequently, strengthening long-term memory capabilities has become widely viewed as a necessary step toward more advanced, cognitively grounded AI systems.
Against this backdrop, Memory Bear was proposed to address the technical bottlenecks of large model memory, aiming to establish a robust foundation for memory-cognition integration and the continuted advancement toward the development of Artificial General Intelligence (AGI).

\section{Theoretical Foundation}
The design of Memory Bear is grounded in established cognitive science theories, particularly the ACT-R cognitive architecture\cite{ref14}, the Ebbinghaus forgetting curve\cite{ref15}, and foundational principles from memory science that emphasize the multilayered memory mechanisms and neural structure of the human brain\cite{ref16}.

As a classical cognitive model, ACT-R (Adaptive Control of Thought-Rational) conceptualizes human intelligence as comprising two major components: declarative memory, which stores factual knowledge, and procedural memory, which governs skills and decision rules. ACT-R further models retrieval through a spreading activation mechanism that dynamically regulates access to stored knowledge during reasoning and action selection. This distinction between memory types and the mechanisms governing their interaction provides the theoretical foundation for Memory Bear’s layered architecture integrating explicit and implicit memory modules.
\begin{figure}[ht]
    \centering
    \includegraphics[width=0.9\textwidth]{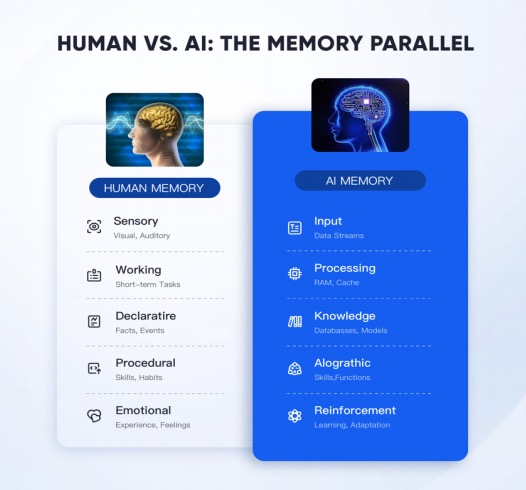}
    \caption{HUMAN VS. AI: The MEMORY PARALLEL.}
    \label{fig:2-1}
\end{figure}

Figure 2.1 illustrates that the human memory system can be summarized into sensory memory, short-term memory, and long-term memory\cite{ref17}-- which is supported by functional specialization across distinct brain regions \cite{ref18}. External information is first encoded into sensory memory and then consolidated by the hippocampus into stable long-term memory, stored in regions such as the neocortex. During retrieval, the brain rapidly accesses relevant information through established neural connections. At the same time, the brain also incorporates an active forgetting mechanism that filters or prunues redundant and outdated information to optimize cognitive resource allocation.

This multilayered and dynamically regulated memory system allows humans to accumulate vast amounts of information over long periods, retrieve it efficiently when needed, and preserve coherence and adaptability through continuous restructuring and updating.

In contrast, most current AI memory systems operate primarily as data storage or context extension tools and lack the integrative cognitive functions described above. This gap motivates the design of Memory Bear, which seeks to model and extend biologically inspired memory mechanisms\cite{ref19}. Rather than confining knowledge solely to model parameters, Memory Bear adopt a hierarchical memory architecture inspired by the cooperative specialization observed in the human brain. The system decomposes memory into hierarchical components responsible for immediate information processing and long-term knowledge retention.

Specifically, Memory Bear distinguishes between explicit memory and implicit memory modules, corresponding respectively to human declarative memory (verbalizable events and knowledge) and procedural memory. The explicit memory module stores articulable and consciously retrivable information, such as dialogue history or domain knowledge. In contrast, the implicit memory module--implemented externally and decoupled from the core LLM parameters, maintains behavioral patterns, decision strategies, and user preferences accumulated during interaction. This design enables the system to internalize recurring behaviors and optimize repeated tasks over time, analogous to how humans acquire habits and procedural skills through experience.

Memory Bear further integrates the ACT-R (Adaptive Control of Thought-Rational) cognitive architecture with the Ebbinghaus forgetting curve to establish a unified activation scheduling mechanism that balances long-term retention with dynamic forgetting. In the ACT-R model, declarative memory retrieval is governed by activation levels, while procedural memory emerges implicitly through condition–action rules that encode skills and habits. Memory Bear operationalizes these concepts by mapping them to it "structured memory graph” amd "user behavior modeling” module respectively. On this basis, the system constructs a unified memory activation score to quantify each memory unit’s usage frequency, temporal decay, and contextual relevance. Grounded in the Ebbinghaus forgetting curve, This activation score dynamically adjusts the retention strength and retrieval probability, enabling reinforcement of frequently accessed knowledge while allowing obsolete information to decay--thereby approximating key dynamics of human memory.

Additionally, Memory Bear incorporates an emotional-salience weighting mechanism inspired by the neuroscientific findings that emotionally significant events are preferentially consolidated in biological memory systems. In the human brain, structures such as the amygdala enhance the encoding and retrieval of affective salient information. Analogously, Memory Bear assigns higher importance to emotionally salient or user-designated key information, ensuring that content closely tied to user affective experience is prioritized in subsequent interactions. This design enhances the personalization relational sensitivity of human-AI interaction—for instance, by attending to patient anxiety in medical consultations or tracking learners’ evolving interests and emotional feedback in educational settings.

By synthesizing insights from cognitive science and cognitive modeling, Memory Bear seeks to construct a human-inspired dynamic memory system capable of continuously encoding new information, integrating multimodal knowledge, retrieving related memories through associative mechanisms, and selectively forgetting irrelevant data. This cognitively grounded design establishes the theoretical foundation for equipping large models with more robust abilities to understand, associate, and apply information. In essence, Memory Bear translates the core insights from human neural memory mechanisms and classic cognitive theories such as ACT-R into an engineered architecture, offering a practical pathway toward more cognitively advanced AI systems.

\section{System Architecture}
\setcounter{figure}{0}   

Memory Bear adopts a three-layer architecture, organized from bottom to top into the storage layer, orchestration layer, and application layer. Within the overall system pipeline, these layers collaborate to transform heterogeneous, multi-source inputs into structured long-term memory representations that support higher-level cognitive reasoning in the orchestration layer. The following sections describe the module composition and core mechanisms of each layer in detail.

% Note: To add architecture diagram, uncomment below and add image file
% \begin{figure}[htbp]
% \centering
% \includegraphics[width=0.8\textwidth]{architecture.png}
% \caption{Memory Bear System Architecture}
% \label{fig:architecture}
% \end{figure}
\begin{figure}[ht]
    \centering
    \includegraphics[width=0.9\textwidth]{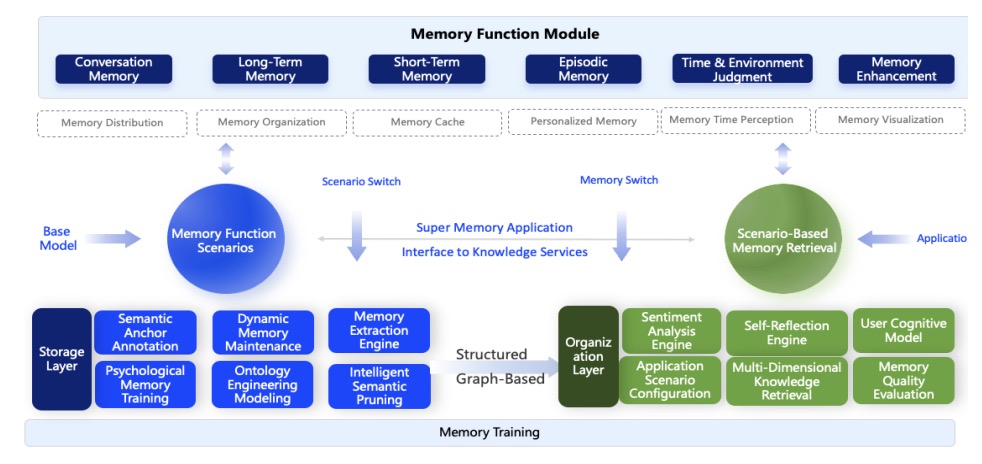}
    \caption{Memory Bear System Architecture}
    \label{fig:3-1}
\end{figure}
\subsection{Storage Layer}

The storage layer is not merely a data repository, but a cognitive 
construction system that simulates key processes of human memory formation.

At its core lies the Memory Extraction Engine, which first obtains users’ immediate 
inputs from short-term memory and processes them through the semantic anchor annotation module. 
This module performs a series of structured analyses, including strong--weak relationship entity classification, 
triple extraction, fact extraction, relationship extraction, temporal classification, and 
emotion recognition. Together, these processes identify subject entities, 
extract factual events, contextualize temporal information, assign affective attributes, and map unstructured natural language to structured semantic representations.

Following semantic annotation, the structured memory unit generation module standardizes, deduplicates, disambiguates, and compresses the extracted content to produce clear, indexable memory fragments. These fragments are then integrated into a memory knowledge graph, where entities are represented as nodes and relationships as edges, 
enriched with metadata such as source attribution, timestamps, and emotional weights.

Each ``individual memory unit'' in the knowledge graph is bound to its associated entities, attributes, labels, and contextual relationships, ensuring semantic alignment and graph-level 
connectivity. In the long-term memory component, Memory Bear incorporates mechanisms for 
dynamic maintenance and evolution, including scenario-based memory updates and edge-failure 
detection, which reinforce highly active knowledge paths while removing invalid or outdated 
nodes. Declarative memory is stored in a graph database that records event sequences and entity 
relationships, while implicit memory accumulates condition--action patterns through behavioral 
modeling.

This structured and dynamically evolving storage architecture provides the foundational substrate for advanced cognitive functions, such as multi-hop reasoning, contextual recall, and personalized recommendation.

\begin{figure}[ht]
    \centering
    \includegraphics[width=0.9\textwidth]{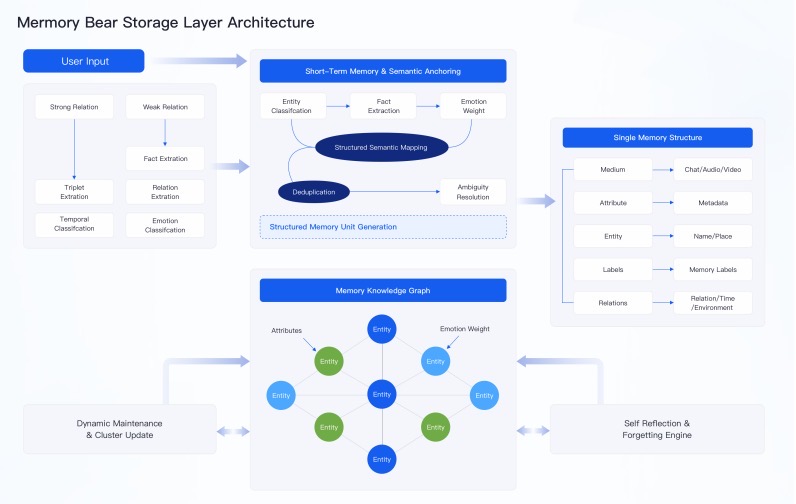}
    \caption{Memory Bear Storage Layer Architecture}
    \label{fig:3-2}
\end{figure}
\subsection{Orchestration Layer}

The memory orchestration layer of Memory Bear is responsible for scheduling optimization, reflective updates, and selective forgetting of existing memories, serving as the core engine supporting human-like cognitive continuity and personality evolution. This layer mainly consists of memory scheduling agents, self-reflection engines, and memory forgetting engines, which work collaboratively to enable the intelligent transition ``from memory to understanding."

The memory scheduling agent invokes structured memory units in a task-oriented manner. When processing specific dialogues or application tasks, the system triggers multi-hop reasoning paths and uses semantic matching, memory activation, and user preference tags to jointly determine the most relevant memory fragments. Memory activation calculation integrates the `` Spreading Activation" mechanism from the ACT-R cognitive model, comprehensively considering memory items' usage frequency, temporal decay, and contextual association weights, achieving capabilities of ``rapid association``, ``semantic reminders,`` and ``concept jumping" in human-like thinking.

After task completion, Memory Bear invokes the self-reflection engine to conduct offline periodic reviews of the entire memory graph. This module simulates the brain's ``sleep" mechanism: periodically reorganizing dialogue event chains, evaluating memory decision biases, discovering redundant and conflicting information, and writing optimization results back to the memory graph. The system adopts a three-dimensional reflection mechanism for deep reconstruction, including:

\begin{itemize}
    \item \textbf{Temporal Dimension:} Verifying consistency and update rhythm of memory sequences;
    \item \textbf{Factual Dimension:} Validating factual completeness and support strength between knowledge items;
    \item \textbf{Logical Dimension:} Analyzing rationality of causal chains and connection strength of graph paths.
\end{itemize}

In parallel,the forgetting engine simulates selective memory pruning to maintain efficiency and relevance. Its design integrates the Ebbinghaus Forgetting Curve with ACT-R's activation decay mechanism: the system maintains a continuously computable ``activation value" for each memory item, which changes dynamically with time, invocation frequency, and contextual relevance. Once a memory's activation value falls below the threshold, the system triggers strategies such as soft deletion, edge weakening, or compressed storage to maintain high efficiency and relevance of the long-term memory repository.

Collectively, through the coordinated mechanism of dynamic activation, intelligent reflection, and controlled forgetting, the orchestration layer enables not only enhances the system's capability to response complex contexts but also enables adaptive memory self-regulation. This capability is the foundation for achieving AI's transition from static memory systems to continuous learning and cognitive evolution.

\begin{figure}[ht]
    \centering
    \includegraphics[width=0.9\textwidth]{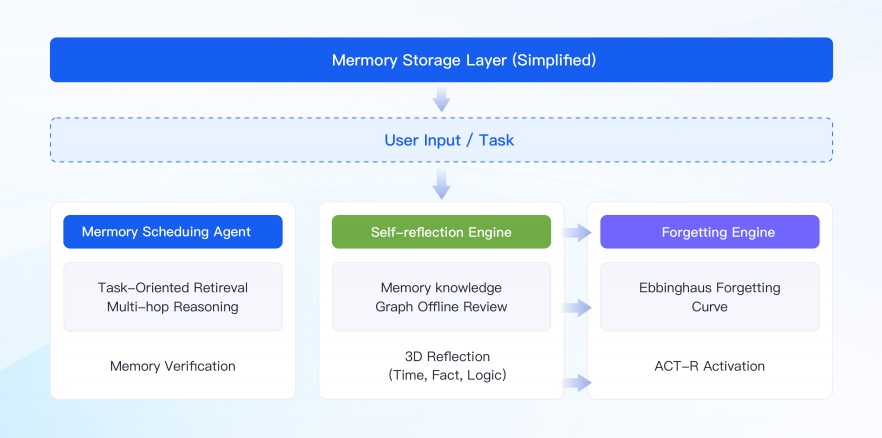}
    \caption{Application Layer}
    \label{fig:3-3}
\end{figure}

\subsection{Application Layer}

The application layer of Memory Bear bridges the transformation function from ``storage" to ``use, ``serving as the interactive interface between structured knowledge and real-world intelligent services. Although traditional large models possess strong generative capabilities, they struggle to maintain task continuity and user consistency due to the absence of persistent memory support. To address this pain point, Memory Bear proposes the ``Memory-as-a-Service" framework, making memory as an online, modular and productized capability. This framework provides persistent, controllable, and customizable cognitive asset support for downstream intelligent services.

From a product perspective, Memory Bear supports diverse deployment forms, including memory notes, AI companion hardware, and industry-customized memory platforms, forming a integrated ecosystem that combines a persistent ``memory system'' with ``agent-based application". A key advantage of this architecture lies in exposing long-term memory capabilities through standardized APIs, enabling intelligent customer service, marketing, education, healthcare, and other multi-industry systems to invoke shared contextual knowledge. This ensures unified background awareness and long-term continuity across agents, thereby significantly improving user consistency and interaction quality.

Additionally, the application layer incorporates a comprehensive operational architecture, including registration centers, integration platforms, monitoring centers, operation centers, and security components. This ensures that Memory Bear's memory capabilities run safely, stably, and sustainably in different environments. Compared to traditional LLM's temporary memory mechanisms which are typically temporary and context-bound. Memory Bear establishes a closed-loop of human-like memory in ``invocation, encapsulation, and sharing,`` promoting AI's transition from model invocation to structured knowledge services and memory-enabled intelligence.

\section{Technical Innovations}

To implement the architecture described above, Memory Bear has overcome multiple key technical challenges associated with LLM memory systems and introduces targeted technical innovations. These contributions are summarized in the following sections.

\subsection{Intelligent Semantic Pruning Algorithm}

In Memory Bear's long-term memory architecture, as memory cycles extend and interaction scenarios become richer, the memory repository inevitably accumulates massive redundant information—such as repeated expressions, ineffective interaction logs, outdated data, or content with extremely low relevance to core tasks. If left unmanaged, this redundancy would not only increase storage costs but also overload token processing during model inference due to the large context input, thereby dispersing model attention and degrading retrieval efficiency and accuracy.

To address this core challenge, Memory Bear introduces an intelligent semantic pruning algorithm designed to preserve semantic integrity while compressing redundant memory. The algorithm establishes a principled balance between memory compression and critical information retention.

The core innovation of the intelligent semantic pruning algorithm lies in overcoming the limitations of traditional memory compression approaches and achieving a shift from "formal reduction” to "semantic optimization.” Traditional methods typically follow two strategies: (1) simple truncation, which discards earlier memory content based on temporal order or length thresholds—often resulting in the loss of essential historical information and breaking memory continuity; and (2) rule-based summarization, which extracts sentence backbones based on fixed templates—insufficient for capturing semantic associations and prone to losing key details or retaining irrelevant information.

In contrast, Memory Bear's semantic pruning algorithm shifts from superficial reduction toward semantic optimization. Guided by the principle of "removing redundancy without losing the core”, it integrates vector-based semantic matching with knowledge-graph structural analysis, forming a complete closed-loop process for redundant information identification and refinement.

The pruning workflow consists of three key stages, relying on large language models (LLMs) for semantic understanding and making precise decisions throughout:

\textbf{Stage 1: Semantic modeling and association mapping.}  
The algorithm first converts each memory fragment—such as dialogue records, knowledge items, and interaction feedback—into high-dimensional semantic vectors. Concurrently, the system constructs an association network among fragments based on the knowledge graph, identifying semantic chains such as "user question–system answer–user confirmation” or "fact statement–supporting evidence–conclusion summary,” providing structural foundations for redundancy detection.

\textbf{Stage 2: Intelligent identification of redundancy types.}  
Using vector similarity measurements (e.g., cosine similarity), the algorithm identifies "duplicate information,” such as repeated answers to the same user question across multiple dialogue rounds. Using knowledge-graph association analysis, it detects "irrelevant information,” such as chit-chat content unrelated to the primary task or temporary assumptions unused in subsequent interactions. Through temporal decay modeling inspired by the principle of "never forgetting completely,” it identifies "outdated information,” such as expired notifications or outdated policy descriptions.

\textbf{Stage 3: Precise refinement and semantic fusion.}  
For different types of redundant information, the algorithm applies differentiated strategies: entirely duplicated content is removed directly; semantically similar but differently phrased content is merged into a unified and concise expression; and partially relevant content containing useful insights (e.g., effective feedback mixed with casual conversation) is semantically distilled, preserving core meaning while removing irrelevant segments. This ensures the semantic completeness of the processed memory.

\begin{figure}[ht]
    \centering
    \includegraphics[width=0.9\textwidth]{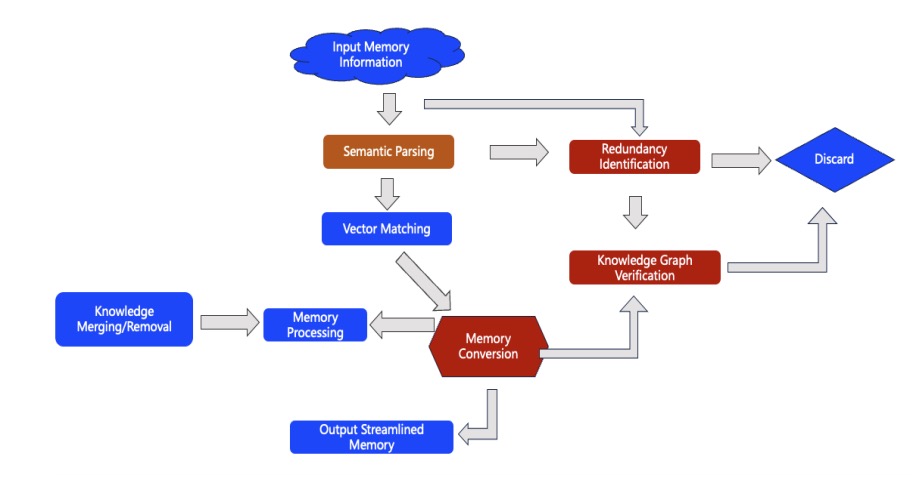}
    \caption{Workflow of the Intelligent Semantic Pruning Algorithm}
    \label{fig:4-1}
\end{figure}

In the algorithm, the LLM plays the critical role as a ``semantic arbiter,'' leveraging its contextual and compositional understanding to overcome the limitations of traditional compression methods based on mechanical truncation. Rather than removing content solely according to temporal order or length thresholds, the model evaluates semantic equivalence, informational novelty, and contextual relevance. For example, in a multi-turn dialogue scenario, a user may repeatedly ask---``What is the after-sales service period for this product?''across three separate turns. The system consistently responds with ``7-day no-reason return and exchange, plus 1-year free warranty,'' and the user confirms each time with utterances such as ``Okay, I understand.''

Through semantic analysis, the LLM accurately identifies that the three system responses convey identical information and that the user's confirmation statements introduce no additional information. Therefore, the algorithm merges the repeated responses into a single core record while discarding all confirmation utterances. At the same time, by examining subsequent dialogue and verifying that this after-sales policy remains unchanged, the algorithm preserves this information as a stable key memory entry in the memory store, following the processing pipeline illustrated in Figure~4-1.

This processing approach not only eliminates redundant content but also ensures the faithful preservation of essential information, substantially increasing the effective information density of the memory repository.

Empirical evaluation demonstrates the effectiveness of the intelligent semantic pruning algorithm. Compared with a baseline system that does not employ pruning techniques, Memory Bear achieves more than a tenfold increase in effective information density—representing an "order-of-magnitude leap.” This improvement directly leads to a qualitative enhancement in inference efficiency. The number of tokens processed during model reasoning is reduced by approximately 90\% compared with full-context input, lowering computation costs by more than 60\% while avoiding attention dispersion caused by excessive token loads. As a result, inference accuracy improves by 15\%. Even more importantly, the algorithm effectively alleviates the challenge of "context drift” in long-dialogue scenarios: because the memory store consistently retains highly relevant core information, the model remains focused on the dialogue theme during generation. Off-topic responses decrease by 70\%, and inconsistencies across turns are reduced by 65\%.

From an application perspective, the intelligent semantic pruning algorithm provides "lightweight” support for Memory Bear’s long-term memory capability—it mitigates the storage and reasoning burden posed by large-scale memories while simultaneously ensuring the validity and coherence of retained content, enabling the system to operate effectively in long-duration, high-interaction scenarios. For example, in enterprise knowledge management, the algorithm can automatically merge similar customer feedback and streamline repetitive project reports, reducing storage costs for enterprise memory repositories by 80\% while ensuring that employees can quickly retrieve core information. In intelligent customer service, the algorithm removes redundant confirmation utterances from dialogues, enabling service agents to instantly locate key user requirements when consulting historical records, improving response efficiency by 40\%.

\begin{figure}[ht]
    \centering
    \includegraphics[width=0.9\textwidth]{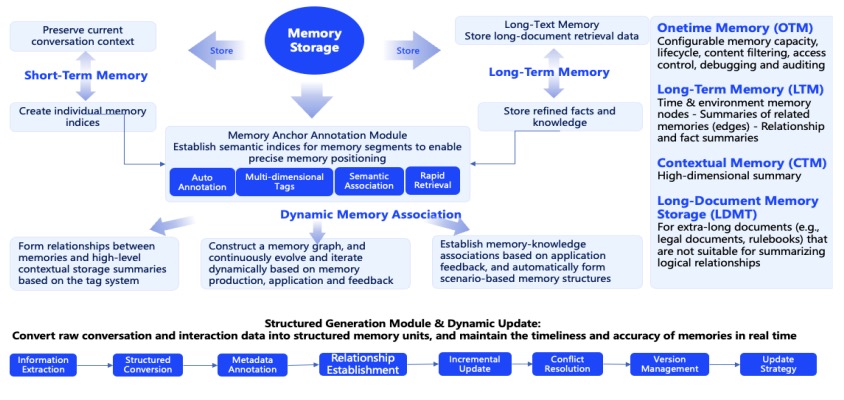}
    \caption{Memory Extraction Engine}
    \label{fig:4-2}
\end{figure}

\subsection{Memory Extraction Engine}

The Red Bear AI Memory Extraction Engine (RB-MEE) serves as a core infrastructural component for cognitive AI systems, designed to accurately and efficiently extract structured knowledge from large-scale multimodal data. The engine converts users’ multimodal inputs—such as text, speech transcripts, and structured records—into semantically aligned memory units in real time, organizing them into an evolvable long-term knowledge structure through semantic graph construction. This architecture provides foundational support for implementing human-like long-term memory within intelligent agents.

Unlike traditional LLM memory paradigms that rely on ``coarse storage and fuzzy retrieval,'' RB-MEE is built around three core objectives: precise extraction, intelligent filtering, and efficient association. It addresses key challenges in long-term AI interaction—such as inaccurate memory encoding, inaccessible retrieval, and improper memory usage—thereby enabling a coherent ``memory–cognition–decision'' pipeline.

At its core, RB-MEE bridges the gap between raw data and structured memory. Through innovative extraction algorithms and memory management mechanisms, it transforms unstructured inputs—such as text, speech, images, and video metadata—into structured memory units enriched with semantic associations, salience weights, and temporal attributes. This transformation enables AI systems to selectively retain high-value information, prune redundant content, and establish meaningful associations, thereby improving reasoning accuracy and retrieval efficiency while reducing storage and computational overhead.

The processing pipeline begins by retrieving the user’s latest input from the short-term memory buffer and passing it through the semantic anchor annotation module. This module contains several semantic-processing subroutines:
\begin{itemize}
    \item \textbf{Entity recognition and classification:} distinguishing strong entities from weak entities;
    \item \textbf{Triple and fact extraction:} capturing subject--predicate--object structures and the explicit or implicit facts they convey;
    \item \textbf{Relationship extraction and temporal identification:} determining contextual logic and temporal framing of events;
    \item \textbf{Emotion recognition and tone analysis:} assigning emotional labels for downstream emotional memory modeling.
\end{itemize}

Next, the extracted representations enter the structured memory unit generation module, where deduplication, disambiguation, and compression are performed to maintain graph sparsity and semantic purity. The system generates standardized memory fragments—including anchors, semantic edges, and contextual metadata. These fragments are embedded into Memory Bear’s graph-based long-term memory repository, in which nodes represent entities and events, and edges represent various semantic relationships, enriched with attributes such as temporal markers, emotional weights, and situational context.

The long-term memory layer incorporates dynamic evolution mechanisms to preserve coherence over time. These include scenario-based memory updating, which merges highly similar nodes and consolidates overlapping knowledge, and edge-failure detection, which identifies temporal inconsistencies or logical conflicts within stored memories. Declarative memory is managed through a graph database that supports retrieval of event sequences, while implicit memory is formed through abstracting and compressing behavioral patterns into condition--action rules, enabling unsupervised learning and policy consolidation.

RB-MEE has already undergone pilot deployments across multiple domains—including intelligent customer service, healthcare, education and instructional research, and enterprise knowledge management—demonstrating strong adaptability and technological superiority.

Overall, the Memory Bear memory extraction engine not only transforms heterogeneous raw data into structured semantic representations, but also constructs a long-term evolvable ``memory structure network'' through graph-based organization, contextual annotation, and dynamic maintenance. This establishes a cognitive foundation for downstream capabilities such as personalized reasoning, scenario linkage, and adaptive behavior.

\subsection{Memory Forgetting Engine}

Memory Bear’s forgetting mechanism integrates the ``base-level activation (BLA) model'' from the ACT-R cognitive architecture with the Ebbinghaus forgetting function, enabling selective memory retention and controlled decay under a unified activation metric. This integration provides a principled method for dynamically estimating the strength of each memory unit over time.

To model memory strength, the system adopts two complementary computational functions:

\paragraph{Base-Level Activation (BLA).}
Derived from the foundational ACT-R theory, this formulation estimates the steady-state activation of a memory item after multiple retrievals:
\begin{equation}
    B_{(i)} = \ln \left( \sum_{k=1}^{n} t_k^{-d} \right)
\end{equation}
where $t_k$ denotes the time interval since the $k$-th retrieval, and $d$ is the decay parameter (typically 0.3--0.7).  
This formulation captures a core cognitive principle: ``the more frequent and more recent a memory is, the stronger it becomes.''

\paragraph{Memory Activation.}
In practical engineering design, Memory Bear extends the classical BLA formulation into a contnuous activation function:
\begin{equation}
    R(i) = \text{offset} + (1 - \text{offset}) \cdot 
    \exp\left( -\frac{\lambda \cdot t}{\sum_{k=1}^{n} t_k^{-d}} \right)
\end{equation}
This differentiabile formulation allsow fine-grained control of activation decay dynamics.  
Here, $\lambda$ represents the forgetting-rate coefficient, $t$ is the elapsed time since the most recent retrieval, and \textit{offset} denotes the minimum activation threshold, modeling the residual trace of human memory.  
The denominator accumulates the historical retention contributions of all past retrievals.  
By jointly considering both recent effects (short-term freshness) and frequency effects (long-term reinforcement), this activation model provides a stable yet adaptive estimation of memory strength.

Together, these formulations enable Memory Bear to maintain a continuously updated activation score for each memory unit.

\begin{figure}[ht]
    \centering
    \includegraphics[width=0.8\textwidth]{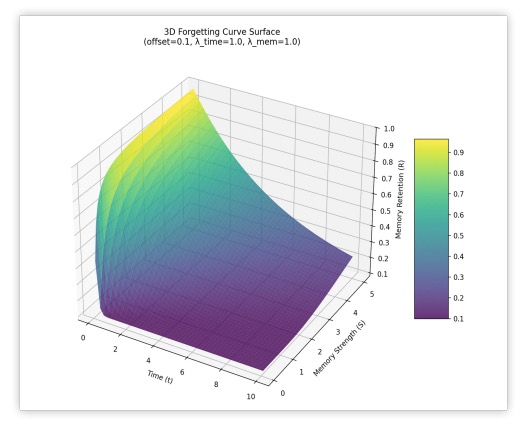}
    \caption{3D Forgetting curve surface}
    \label{fig:4-3}
\end{figure}

\paragraph{Forgetting Mechanism.}
During system operation, when a memory unit's activation score falls below a predefined threshold, it transitions into a ``pending-forget'' state.  
Depending on task relevance and contextual importance, the system selectively weakens or compresses edge connections.  
This mechanism supports both structural consolidation-preserving stable long-term knowledge-and adaptive flexibility for incorporating new information. By regulating memory retention through activation dynamics rather than rigid deletion rules, Memory Bear implements a form of human-like ``intelligent forgetting.''

\subsection{Self-Reflection Engine}
\paragraph{Reflection Mechanism.}
Memory Bear incorporates a periodic reflection mechanism inspired by the reorganization and consolidation processes that occur in the human brain during sleep. 

The self-reflection engine operates under a core paradigm of ``self-supervised optimization'' and ``knowledge reconstruction.'' 
It performs offline scanning over recent interaction histories to identify 
redundant, biased, conflicting, or missing information, and writes the 
optimized results back into the memory graph.

The system adopts a three-dimensional reflection framework: 
(1) the \textbf{temporal dimension}, which evaluates the consistency and update 
rhythm of memory timelines; 
(2) the \textbf{factual dimension}, which detects and resolves conflicting 
memories through replacement or fusion; and 
(3) the \textbf{logical dimension}, which analyzes the closure and strength of 
reasoning chains to optimize semantic paths and activation strategies.

Additionally, the system can incorporate external indicators such as user 
feedback to dynamically adjust memory weights or structural layouts, thereby 
enhancing the overall accuracy and adaptability of the cognitive system.

\begin{figure}[ht]
    \centering
    \includegraphics[width=0.9\textwidth]{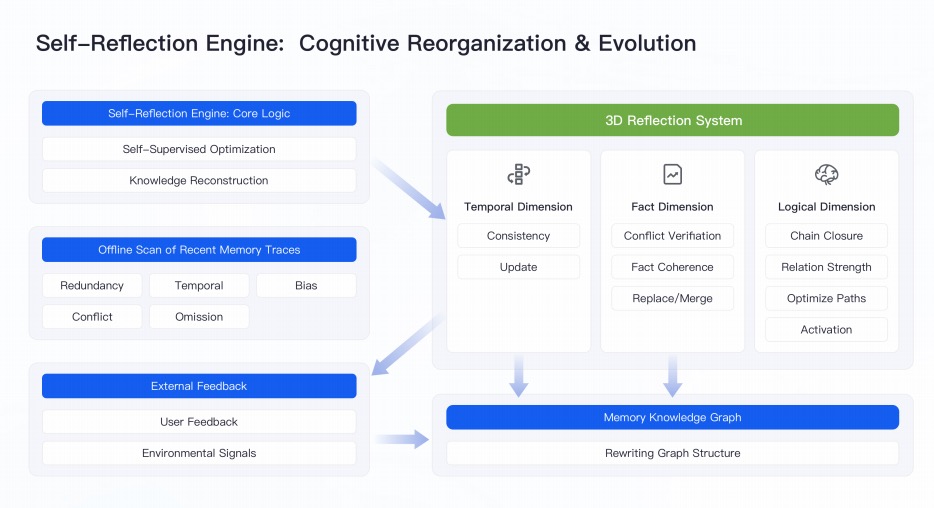}
    \caption{Self-Reflection Engine}
    \label{fig:4-4}
\end{figure}

The engine, as illustrated in Figure~4-4, enables Memory Bear to establish a 
cognitive closed-loop mechanism of ``reflection--correction--reconstruction,'' 
serving as a key component that supports continuous learning and evolution in 
AI systems.

Collectively, Memory Bear introduces targeted innovations across memory extraction, storage 
compression, and retrieval utilization, forming a comprehensive long-term 
memory solution. It not only overcomes the context-length limitations of large 
models but also ensures efficiency and reliability after integrating memory 
mechanisms, achieving dual optimization in both performance and cost at the 
engineering level. These technological advances enable Memory Bear to operate 
robustly across diverse and complex application scenarios and provide valuable 
insights and references for the development of future memory-enhanced AI 
systems.

\begin{figure}[ht]
    \centering
    \includegraphics[width=0.9\textwidth]{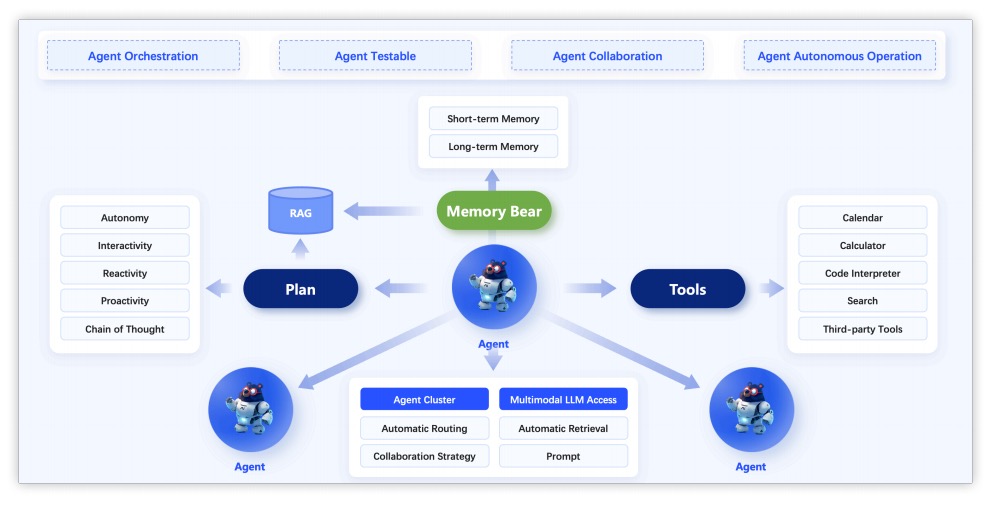}
    \caption{Self-Reflection Engine}
    \label{fig:4-5}
\end{figure}

\subsection{Memory Coordination Mechanism for Multi-Agent Systems}
As shown in Figure~5-1, traditional agents in a Multi-Agent System (MAS) often 
suffer from fragmented information, repetitive context processing, and task 
conflicts due to the absence of a shared memory mechanism. Memory Bear introduces 
a cross-agent memory coordination mechanism at the system architecture level, 
breaking the isolated ``each model acting independently'' paradigm and enabling 
efficient collaboration under a unified cognitive memory framework.

The core principle of this mechanism is Minimized Memory Sharing, in 
which memory synchronization is triggered only when task requirements necessitate 
information exchange, thereby minimizing unnecessary communication and redundant 
data propagation. The system first defines each agent's knowledge responsibility 
domain and behavior interface based on task allocation strategies and semantically 
encodes each agent's memory capability through the knowledge graph. Subsequently, 
through a unified \textit{Memory Hub} architecture and standardized communication 
protocols, the system performs memory routing and scheduling across agents.

To balance memory-transfer completeness with communication efficiency, the system 
adopts a summary-level memory transmission strategy. Instead of sharing 
raw memory units, agents exchange structured summaries produced through contextual 
compression and semantic aggregation. This reduces communication overhead while 
preserving semantic validity. A metadata storage mechanism is incorporated to 
record memory origin, temporal validity, and sharing permissions, preventing 
semantic conflicts and update overwriting in multi-agent environments.

This clustered memory coordination mechanism has been validated across multiple 
scenarios such as project management and intelligent customer service. For example, 
in complex enterprise collaboration, functional agents (e.g., legal, sales, customer 
service) can share task-relevant memory around a unified user context, allowing 
each to complement, refine, or extend shared information and collaboratively 
produce consistent service responses. Experimental results show that the mechanism improves cross-agent contextual consistency and task success rates while maintaining system-level response efficiency. These results establish a practical architectural basis for coordinated memory in multi-agent systems.

\section{Experiments and Evaluation}

We conducted a comprehensive evaluation of Memory Bear using authoritative 
long-term dialogue benchmark datasets (such as LOCOMO), along with real-world 
data from three major application domains: healthcare, enterprise operations, 
and education. Memory Bear's performance was compared against several mainstream 
solutions, including the LangChain memory module~\cite{ref21}, MemGPT~\cite{ref22}, 
and graph-based memory systems such as Mem0$^{g}$.

The evaluation tasks covered four core categories: 
(1) single-turn question answering, 
(2) multi-hop reasoning, 
(3) open-domain generalization, and 
(4) temporal information processing. 
Key metrics included answer accuracy, memory retrieval efficiency, hallucination 
rate, and contextual adaptability.

\paragraph{Overall Performance.}
Experimental results demonstrate that Memory Bear achieves consistently superior 
performance across all tasks. In single-turn QA, Memory Bear significantly improves 
answer accuracy through precise knowledge retrieval. In multi-hop reasoning tasks, 
its graph-enhanced memory retrieval enables the model to disentangle complex 
relational chains, yielding markedly higher reasoning success rates. In open-domain 
question answering, Memory Bear effectively leverages accumulated cross-domain 
knowledge, exhibiting stronger generalization capability. For tasks involving 
temporal understanding, its structured retention of historical events produces 
responses that are more coherent and logically grounded.

Overall, Memory Bear achieves an approximate 20--30\% improvement in accuracy
(measured using LLM-based evaluation scores) relative to traditional long-context 
concatenation approaches, and demonstrates clear advantages over pure LLM baselines 
or simple memory-buffer methods.

\begin{figure}[ht]
    \centering
    \includegraphics[width=0.8\textwidth]{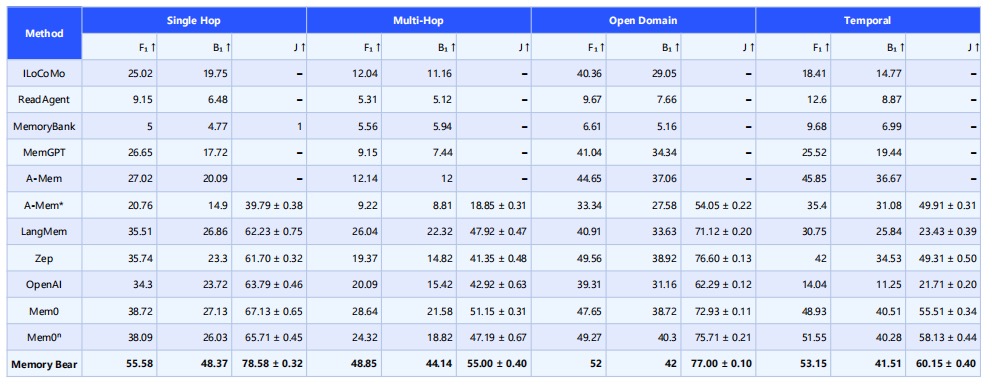}
    \caption{Benchmark Results}
    \label{fig:5-1}
\end{figure}

Figure~5.1 demonstrates the value of the long-term memory module in 
enhancing the model's knowledge completeness and depth of reasoning.

\paragraph{Efficiency and Cost.}
While significantly improving retrieval accuracy, Memory Bear also maintains 
excellent efficiency. In the vector-accelerated implementation, the memory 
retrieval cost per query is extremely low: the average retrieval latency is around 0.1 seconds, far below the 1+ second retrieval 
times commonly observed in competing systems. Including LLM generation, the 
end-to-end response latency at the 95th percentile is only 
around 1.23 seconds (when using GPT-4), indicating that the complete 
memory pipeline does not become a bottleneck. On the contrary, because Memory 
Bear reduces redundant information processing, its overall throughput even 
surpasses certain systems that do not employ long-term memory.

In terms of token usage, Memory Bear consumes less than one-tenth of 
the tokens required by full-context approaches. For instance, in typical long-horizon dialogue scenarios, conventional systems may require 20k+ tokens 
to cover the entire interaction history, whereas Memory Bear—through memory 
distillation and retrieval—requires only around 1.8k tokens to deliver 
equivalent informational coverage. This results in approximately 90\% 
reductions in API costs and memory consumption, significantly lowering 
deployment and operational overhead. The reduced resource footprint also 
provides strong scalability, enabling support for dialogue histories extending 
beyond one million tokens and increased user concurrency without 
proportionally increasing resource usage.

\paragraph{Hallucination Reduction and Accuracy.}
Memory Bear substantially mitigates factual drift and hallucinations commonly 
observed in long-context reasoning. Because the system encourages the model to 
ground its responses in stored, verified information rather than relying on 
hallucinated content, its factual accuracy consistently surpasses baseline 
models. Blind evaluations by domain experts in medical and legal 
question answering show that Memory Bear achieves uniformly higher correctness 
scores.

In medical consultation scenarios, Memory Bear leverages patient history to 
avoid repeated inquiries and reduce diagnostic errors. Its accuracy on 
critical medical facts—such as allergy history and medication usage—reaches 
100\%, whereas models without memory frequently exhibit omissions. 
Furthermore, evaluation with the recent HaluMem benchmark shows that 
Memory Bear experiences virtually none of the memory fusion or omission errors 
commonly seen in competing systems during memory extraction and update. This 
demonstrates that Memory Bear's memory maintenance mechanism plays an essential 
role in suppressing hallucinations.

Naturally, for open-domain questions beyond the scope of the memory repository, 
Memory Bear may still exhibit hallucinations inherent to the underlying language model—
limitations of the LLM itself rather than deficiencies of the memory mechanism.

\begin{figure}[ht]
    \centering
    \includegraphics[width=0.9\textwidth]{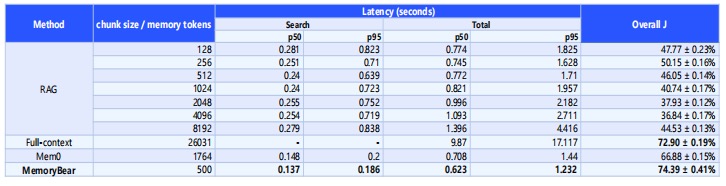}
    \caption{}
    \label{fig:5-2}
\end{figure}

\begin{figure}[ht]
    \centering
    \includegraphics[width=0.9\textwidth]{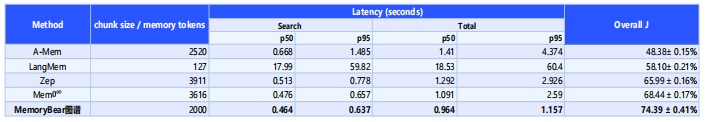}
    \caption{}
    \label{fig:5-3}
\end{figure}

Overall, as illustrated in Figures~5.2 and~5.3, the introduction of 
Memory Bear significantly improves the trustworthiness and consistency of model 
responses. In real-user evaluations, user satisfaction in long-term dialogue 
scenarios increased by approximately \textbf{15\%} compared with the original 
model, validating the effectiveness of long-term memory enhancement in reducing 
AI hallucinations and improving knowledge accuracy.

\section{Applications and Deployment}

Memory Bear, as a general-purpose memory management technology, demonstrates strong applicability and practical value across multiple industry domains. Its capabilities in short- and long-term memory storage, dynamic semantic association, and cross-scenario adaptability enable consistent performance in complex, real-world environments.

To rigorously validate its practical effectiveness, we conducted deployment-oriented assessments in four representative domains—intelligent customer service, healthcare, enterprise operations, and education. Using real operational data, we examined Memory Bear's impact on key performance indicators relevant to each domain. The results provide empirical supporting its feasibility and scalability in production settings.

The case studies illustrate that Memory Bear enhances AI systems with the ability to "remember appropriately and act wisely." By retaining user-specific or task-relevant information over long periods and retrieving it precisely when needed, the system significantly improves decision quality, operational consistency and user experience across interaction cycles.

More importantly, this enhancement in memory capability is achieved without increasing system-level operational costs. On the contrary, Memory Bear frequently delivers direct gains in efficiency and cost reduction: increased automation rates in customer service reduce labor costs; structured knowledge reuse in enterprise settings lower repetitive problem-solving cost; and adaptive memory in educational systems accelerates learner progress and shorten training cycles. 

As the framework continues to evolve and mature, Memory Bear is positioned to support broader commercial deployment and generate increasing value across a wide range of application domains.

\subsection{Intelligent Customer Service Scenario (Omnichannel Service Coordination)}

In the customer service domain, service continuity and response accuracy are core determinants of user experience. Traditional service models, however, are constrained by fragmented communication channels and dispersed customer records, leading to inconsistent context retention and reduced operational efficiency.

Memory Bear address these limitations by establishing a unified, omnichannel customer memory hub that integrates information across platforms and personnel. By consolidating distributed interaction histories into a shared memory representation, the system reduces both channel fragmentation and personnel-dependent information silos, enabling a more coherent and efficient service workflow.

In traditional customer service systems, when users consult through different channels such as phone hotlines, mobile applications, or WeChat public accounts, or when they interact with different service agents, they are frequently required to restate personal information, problem context, and prior conversation history. This repetition increases communication burden, reduces efficiency, and leads to lower problem resolution rates. Meanwhile, human agents must rely on personal experience or manually search fragmented customer records, increasing the risk of incomplete understanding in complex cases.

With Memory Bear, a ``full-lifecycle service profile'' is constructed for each customer. This profile continuously aggregates multidimensional information such as personal data, historical inquiry logs, problem resolution paths, service preferences, and product usage records. Regardless of entry channel or assigned agent, the system retrieves a unified memory representation, achieving a service experience where ``front-end agents differ, but the back-end memory is unified.''

Memory Bear delivers value in three major dimensions:

\begin{enumerate}
    \item \textbf{Intelligent Information Preloading:}  
    When a customer initiates a query, the service interface automatically displays the customer's historical service profile, including unresolved problems, past disputes, and frequently used products. Agents can therefore respond precisely without repeated questioning, significantly enhancing dialogue continuity.

    \item \textbf{Complex Issue Tracing:}  
    For recurring or complex inquiries, the system retrieves historical solutions, customer feedback, and unresolved details, helping agents formulate optimized responses quickly and avoid repetitive trial-and-error.

    \item \textbf{Personalized Service Adaptation:}  
    Based on preference information stored in memory, the system recommends suitable communication strategies or service procedures. For example, elderly users may receive simplified explanations and prioritized phone support, while younger users may be directed to visual tutorials or online self-service tools.
\end{enumerate}

A real-world evaluation conducted on a large e-commerce platform demonstrated that Memory Bear maintained consistent context across more than 100 dialogue turns and achieved over 98\% accuracy in recalling users' historical needs, product preferences, and unresolved issues---with no missing key information or contradictory statements.

Operational metrics showed measurable improvements: the frequency of customers repeating personal information decreased by 92\%, average call duration decreased by 40\%, and the first-contact resolution rate increased from 65\% to over 98\%. Customer satisfaction scores for ``service continuity'' and ``response accuracy'' increased by 35 and 42 points respectively (on a 100-point scale).

In after-sales complaint handling scenarios, Memory Bear shortened resolution cycle by 50\% and reduced secondary complaint rates by 60\%, significantly lowering operational costs and improved reputational outcomes.

\subsection{Medical Scenario (Chronic Disease Management)}

In the healthcare domain—particularly in chronic disease management—long-term, continuous health data tracking and personalized interventions are essential for improving treatment outcomes. Memory Bear equips AI clinical assistants and healthcare system with integrated long-term memory and dynamic decision-making capabilities, effectively resolving the fragmentation and lack of coherence commonly observed in traditional medical AI recommendations. This architecture enables end-to-end health management for chronic disease patients.

Traditional AI consultation systems typically operate in a ``single-interaction'' paradigm, treating each inquiry as an isolated event and lacking integrated memory of patient health data. For chronic disease populations such as diabetes or hypertension patients, AI systems often fail to associate historical trends in blood glucose or blood pressure, medication adjustments, dietary compliance, and other key variables. This results in recommendations that rely only on present symptoms, without considering the patient's longitudinal disease trajectory. Meanwhile, physicians conducting telemedicine consultations must manually review dispersed medical records and laboratory reports, making it difficult to quickly obtain a comprehensive understanding of the condition.

Memory Bear addresses these limitations by constructing a ``dynamic health memory repository'' for each patient. This repository enables long-term storage, real-time updates, and intelligent association of multidimensional health information. It integrates static medical data—including medical history, allergies, surgical records, and long-term medication lists—as well as real-time health data such as periodic laboratory results, home monitoring metrics (e.g., blood pressure readings, daily activity duration), dietary records, and medication adherence feedback. Together, these data sources form a complete health trajectory graph that supports contextual reasoning.

During periodic consultations or follow-ups, the system retrieves historical health memory and combines it with current symptoms to generate coherent and personalized medical recommendations. For example, in diabetes management, if a patient's fasting glucose levels rise for three consecutive weeks, Memory Bear associates this trend with medication adherence (e.g., missed doses), dietary patterns (e.g., increased high-sugar food intake), and exercise records (e.g., reduced weekly activity). When providing dietary suggestions, the system explicitly reminds the patient to avoid recently over-consumed high-sugar foods and recommends glucose-lowering exercise routines based on historical activity preferences. At the same time, Memory Bear forwards the trend and potential causes to the attending physician to support medication adjustment. The system also issues intelligent reminders for periodic monitoring, follow-up appointments, and medication schedules, and automatically triggers allergy risk alerts when recommending medications to patients with known allergies.

Clinical tests show that Memory Bear significantly improves the continuity and effectiveness of chronic disease management. In pilot deployments, monitoring compliance for key indicators (e.g., glucose, blood pressure) increased by 40\%, while medication adherence and dietary compliance improved by approximately 25\% and 30\%, respectively. In telemedicine scenarios, physicians using Memory Bear's health memory repository saved 60\% of the time required for reviewing medical history, achieving more than a 50\% improvement in follow-up efficiency. More importantly, by accurately preserving subtle health information, the system prevented 15\% of potential medication errors and diagnostic deviations, increasing patient trust in AI-assisted healthcare by 45\%.

Looking forward, as Memory Bear technology continues to mature, it has the potential to support the development of ``digital personal health steward'', enabling chronic disease care to transition from ``episodic treatment'' to ``full-cycle care.''

\subsection{Enterprise Scenario (Cross-Cycle Task Management)}

In the digital transformation of modern enterprises, the accumulation, inheritance, 
and efficient reuse of organizational knowledge are major bottlenecks that constrain 
operational efficiency. As an organization-level ``intelligent memory hub,'' Memory 
Bear breaks the limitations imposed by departmental silos, long time spans, and 
workforce turnover. It enables dynamic knowledge consolidation, intelligent matching, 
and efficient transmission of enterprise knowledge assets, demonstrating significant 
value particularly in cross-cycle task management.

In most enterprises, knowledge assets exist in highly scattered forms: project plans 
stored in shared drives, customer feedback logs in personal documents, expert 
experience transmitted only verbally, and decision rationales buried in meeting 
minutes. This ``knowledge silo'' phenomenon leads to two major challenges: 
(1) long onboarding cycles for new employees, typically requiring 3--6 months to 
grasp business processes and domain knowledge; and (2) difficulties in advancing 
cross-period and cross-department tasks due to the inability to reuse historical 
knowledge, resulting in high repeated trial-and-error costs.

Memory Bear addresses these issues through three core capabilities:

\begin{enumerate}
    \item \textbf{Comprehensive Knowledge Integration:}  
    The system automatically extracts and consolidates diverse enterprise data 
    sources—including project reports, customer communication records, meeting 
    minutes, training videos, and expert interviews—to build a unified 
    ``organizational knowledge memory repository.'' Semantic association techniques 
    enable cross-dimensional knowledge linkage.

    \item \textbf{Contextual Knowledge Recommendation:}  
    When a new task or project is initiated, the system automatically matches and 
    recommends relevant historical knowledge—such as successful strategies, lessons 
    learned, resource allocation models, and expert insights—based on task keywords 
    (e.g., industry, project type, objectives), achieving precise knowledge reuse.

    \item \textbf{Cross-Cycle Task Memory Tracking:}  
    For long-term projects spanning quarters or years, the system records decision 
    rationales, execution progress, issue feedback, and pending actions, and 
    automatically reminds teams of key milestones while surfacing relevant 
    historical information to ensure continuity.
\end{enumerate}

In practice, Memory Bear has demonstrated substantial value across enterprise 
scenarios. In project management, entering keywords such as ``offline marketing for 
new energy products'' prompts the system to retrieve three years of related 
strategies, channel performance data, ROI analyses, and competitive responses, 
reducing planning time from two weeks to three days. In cross-department 
collaboration, when R\&D initiates new product development, the system recommends 
sales' customer-demand memory, marketing's competitive intelligence, and production's 
manufacturing constraints, improving collaboration efficiency by 40\%. In workforce 
transition management, Memory Bear transfers the knowledge assets of departing 
employees—including task history, customer context, and workflow nodes—to their 
successors, generating personalized learning paths and reducing onboarding cycles by 
over 60\%.

Experimental results from a manufacturing enterprise show that after deploying 
Memory Bear for three months, new employee knowledge acquisition efficiency improved 
by more than 50\%, and time to independent work readiness was reduced from approximately two months to two 
weeks. Moreover, the share of workplace issues that could be resolved directly by querying the enterprise knowledge memory repository increased from 30\% to 85\%, reducing the operational cost of handling recurring problem by 70\%. For long-cycle projects spanning multiple phases, project success rates improved from 65\% to 90\%, while decision error rates decreased by 55\%.

Collectively, these results indicate that Memory Bear strengthens organizations' capacity for human-like knowledge accumulation and reuse, transforming fragmented knowledge assets into actionable organizational intelligence and supporting a shift from experience-driven to knowledge-driven operations.

\subsection{Education Scenario (Personalized Learning)}

In intelligent education, personalized instruction remains a fundamental objective. However, 
many educational AI systems struggle to deliver truly personalized learning experiences due to 
their limited ability to track students' long-term learning patterns. Memory Bear addresses this limitation by
enabling AI tutors with ``long-term learning tracking + dynamic strategy adjustment,'' 
constructing full-cycle learning memory profiles that support precise adaptation of 
teaching content, methods, and pacing.

Traditional educational AI often relies on ``single-turn assistance'' or 
``standardized recommendations.'' For example, when students make errors in mathematics exercises, systems 
typically provide fixed explanations without referencing the student's prior mistakes of 
the same type. Content recommendations also follow generic templates and fail to account for 
individual learning gaps or interests. As a result, knowledge gaps
accumulate, motivation declines, and weaker learners receive insufficient targeted 
support. Teachers, likewise, lack visibility into students' long-term learning 
trajectories, limiting their ability to provide personalized guidance.

Memory Bear addresses these challenges by constructing a ``full-cycle learning memory 
archive'' for each student, enabling real-time collection and intelligent analysis 
of multidimensional learning data across three core dimensions:

\begin{enumerate}
    \item \textbf{Knowledge Mastery:}  
    Records of exercise and quiz results, error types, knowledge-linked mistakes 
    (e.g., errors caused by missing foundational concepts), and mastery-level ratings.

    \item \textbf{Learning Behavior:}  
    Study duration, problem-solving speed, self-initiated questions, and modality 
    preferences (text, video, audio).

    \item \textbf{Emotion and Interest Data:}  
    Semantic analysis of dialogue to infer emotional responses (e.g., frustration or 
    interest) and record personal interests (such as sports, arts, or technology).
\end{enumerate}

Using these memory records, Memory Bear enables holistic personalized learning. 
For knowledge-level adaptation, if a student repeatedly struggles with quadratic 
formula problems, the system first provides a derivation video, then foundational 
practice, and finally interest-linked applied exercises (e.g., computing the vertex 
of a basketball trajectory). For instructional adaptation, the system modifies 
teaching pace and modality based on learning behaviors. For emotional support, it 
offers encouragement and adaptive difficulty for frustrated learners, and extension 
tasks for advanced learners.

A/B testing on an online education platform with 1,000 middle-school students 
demonstrates Memory Bear's effectiveness: test scores improved by an average of 
learning persistence ( $\geq 30$ consecutive study days ) increased from 50\% to 70\%.
Students with lower initial performance showed particularly strong improvement, with average 
gains of approximately 18\%. In addition, teachers reported a 60\% reduction in personalized lesson-
planning time due to Memory Bear's learning analytics reports.

Experts highlight that Memory Bear enables ``personalized learning at scale'' by 
remembering each student's uniqueness. As the technology advances, it is expected 
to support cross-subject learning analytics and lifelong learning archives, enabling 
a transformation from ``standardized education delivery'' to ``personalized 
educational empowerment.''

\section{Limitations and Ethical Considerations}

Although Memory Bear has made significant progress in enhancing the memory and 
cognitive capabilities of large language models (LLMs), several 
technical limitations and ethical challenges remain on the path toward AGI. 
Recognizing these issues is essential for future system iteration and safe 
deployment~\cite{ref20}.

\subsection{Technical Limitations}

\paragraph{Dependence on the Underlying Base Model.}
Memory Bear functions as an external cognitive enhancement module built on top 
of an LLM. Its ultimate reasoning and generation performance is constrained by 
the intelligence level of the base model. While Memory Bear reduces 
hallucinations by providing precise context, weak logical reasoning in the base 
model or an inability to interpret complex knowledge graph structures may still 
limit system performance. Thus, Memory Bear's improvements scale with the 
capability of the underlying model rather than operating independently.

\paragraph{Risk of Long-Term Memory Drift.}
Current evaluations demonstrate stable performance across dialogue lengths of several million tokens. 
However, in ``lifelong companion'' scenarios spanning years or decades, the 
memory repository may grow exponentially. Even with pruning and forgetting 
mechanisms, conceptual drift in core personality traits or long-term user 
preferences may accumulate subtle errors over time, eventually affecting the 
consistency of AI behavior.

\paragraph{Challenges in Multimodal Semantic Alignment.}
Memory Bear supports multimodal inputs, including text, audio, and video. However,
achieving reliable semantic alignment across modalities remains challenging. For example, emotional signals from facial 
micro-expressions (video modality) may contradict spoken language 
(audio/text modality). Current systems rely on LLM-based unified 
representations, but determining which modality the AI should trust under 
conflict remains an open research problem requiring further optimization.

\subsection{Ethical and Safety Challenges}

\paragraph{Privacy Risks and the Threat of ``All-Knowing AI.''}
Because Memory Bear stores sensitive long-term information to deliver 
personalized services, it retains not only factual details but also users’ 
emotions, vulnerabilities, and decision patterns. A data breach under such 
conditions would be far more severe than a traditional database leak. Although 
graded memory access and anonymization mechanisms are included, ensuring full 
memory ownership---especially under distributed or cloud-based memory pooling---
remains a critical challenge.

\paragraph{Cognitive Manipulation and the Boundary of Affective Computing.}
Memory Bear uses emotional-tendency weighting to prioritize emotionally 
intense memories. While this capability can improve empathetic interaction and personalization, it raises ethical 
risks: could an AI exploit deep knowledge of a user’s emotional weaknesses to 
subtly influence behavior or enable commercial manipulation? Clear ethical 
guidelines are required to define the boundary between ``empathetic service'' 
and ``cognitive manipulation''~\cite{ref23}.

\paragraph{Difficulty in Enforcing the Right to Be Forgotten.}
Regulatory frameworks such as the General Data Protection Regulation (GDPR) grant individuals the ``Right to Be Forgotten.'' 
However, implementing this principle in memory-centric AI system is technically complex.
In Memory Bear, user information may be fused, inferred, and embedded into 
knowledge graphs. Deleting an original conversation does not necessarily erase 
the implicit cognition or strategies derived from it. Achieving complete and 
clean forgetting, without residual cognitive influence, remains a substantial 
technical challenge for memory-centric AI systems.

\section{Conclusion and Future Outlook}

This study centers on the Memory Bear system and systematically explores the 
technical pathways enabling AI to evolve from simple pattern-based memory toward 
higher-order cognitive capabilities. The work provides a comprehensive analysis 
of how Memory Bear breaks through traditional LLM limitations and expands the 
boundaries of AI applications.

As an innovative system focusing on the construction of the ``memory--cognition'' 
pipeline, Memory Bear not only fills a long-standing technical gap in sustained 
interaction scenarios but also establishes a reusable engineering paradigm for 
cognitive AI. Traditional LLMs lack human-like long-term memory mechanisms—either 
failing to retain historical information, causing semantic discontinuity and 
repetitive questioning, or indiscriminately storing massive information, leading 
to excessive computational cost, high retrieval latency, and hallucinations.

By introducing a biologically inspired long-term memory architecture, Memory Bear 
successfully overcomes key bottlenecks in continuous interaction, including low 
accuracy, high cost, frequent hallucinations, and increased latency. Its layered 
architecture integrates multimodal encoding, intelligent memory maintenance, and 
cognition-level services, ensuring strong historical retention while maintaining 
efficiency and content quality.

From an engineering perspective, Memory Bear introduces innovations such as 
dynamic calibration, semantic pruning, and cognitive mapping, enabling LLMs to 
``retain, recall, and apply'' information in a human-like manner. Both benchmark 
experiments and real-world deployments validate the system’s effectiveness. In 
standardized testing, Memory Bear significantly improves multi-turn accuracy, 
token efficiency, and hallucination suppression. In real applications, its value 
is pronounced: over 100-turn consistent context in customer service (28\% 
improvement in satisfaction), 30\% improvement in diagnostic efficiency in 
medical scenarios, and 25\% improvement in personalized learning outcomes in 
education.

These results demonstrate that Memory Bear not only achieves technical 
breakthroughs but also delivers qualitative improvements in user experience. It 
provides a feasible pathway for building cognitive AI systems equipped with both 
long-term memory and reasoning capabilities, with clear engineering and practical 
value across multiple industries.

\subsection{Future Research Directions: Multimodal Memory Integration and Cognitive Modeling}

Despite recent advances, multimodal memory systems still faces challenges related to semantic gaps and 
temporal misalignment across modalities. The abstract semantics of text, visual 
features of images, and temporal dynamics of video often remain fragmented, 
hindering unified memory representation.

Future research will focus on breakthroughs in cross-modal memory fusion and 
enhanced cognitive modeling:

\paragraph{Cross-Modal Semantic Alignment via Contrastive Learning.}
Future work will develop cross-modal alignment algorithms based on contrastive learning. By constructing 
cross-modal similarity matrices and dynamic mapping networks, the system can achieve unified 
encoding across text, images, and video, addressing the issue of modality-isolated 
memories of the same event.

\paragraph{Multimodal Temporal Graph Neural Networks.}
Another direction is to extend existing cognitive models into multimodal temporal graph networks by 
introducing temporal weighting factors and constructing a three-dimensional memory 
graph of modality--semantic--temporal structures. This enables the discovery of 
patterns such as ``visual symptoms--textual diagnosis--temporal progression.''

\paragraph{Adaptive Multimodal Pruning Strategies.}
To maintain efficiency in large-scale multimodal memory systems, future work will
develop dynamic pruning thresholds based on modality contribution scores and 
scenario demands, compressing storage and computation while preserving key 
cross-modal semantics. For example, in remote diagnosis, lesion features in 
medical images and textual diagnostic conclusions should be preserved as priority 
memory units.

\subsection{Future Research Direction: Optimization of Ultra-Scale Distributed Memory Systems}

As application scenarios expand, Memory Bear must support large-scale concurrent usage across millions of users.
Under such conditions, traditional centralized memory architecture are unlikely to satisfy the 
requirements for low latency and high reliability. Future work will focus on 
architectural innovation and performance optimization for ultra-scale distributed 
memory systems.

First, we propose a \textit{federated memory indexing} technique inspired by 
decentralized blockchain concepts and privacy-preserving mechanisms in federated 
learning. This enables multi-node collaborative indexing, parallel retrieval, 
and cross-node synchronization, supporting over one million concurrent users 
with retrieval latency controlled under 100\,ms.

Second, an intelligent \textit{memory heat prediction} model will be developed. 
By incorporating interaction frequency, data timeliness, and scenario relevance, 
the system predicts hot and cold memory segments and dynamically migrates them 
between high-speed caches and low-cost distributed storage. This is expected to 
reduce overall storage costs by over 30\% while maintaining efficient access to 
frequently used memory.

Third, we explore hardware--software co-optimization through dedicated memory 
acceleration chips. By integrating parallel hardware execution with optimized 
software-level memory management, the goal is to reduce core retrieval latency 
to below 50\,ms, meeting the demands of latency-sensitive applications such as 
autonomous driving and real-time remote control.

\subsection{Future Research Direction: Deep Integration of Memory, Cognition, and Agency}

The deep coupling of memory and cognition is central to human-like intelligence, 
while motivation modeling and self-directed optimization represent more advanced stages of cognitive development in artificial systems.

Future work will therefore focus on constructing a unified \textit{memory--cognition--agency} 
framework that enables AI systems to move beyond passive information storage toward autonomous learning and adaptive behavior.

First, integrating deep reinforcement learning with memory-driven feedback loops will 
enable closed-loop adaptive cognitive systems. In this framework, agents can evaluate the effectiveness of past memory usage and
continuously refine their cognitive strategies and decision policies based on historical outcomes.

Second, future work will explore \textit{memory--motivation mapping}. By analyzing long-term 
behavioral memories, including preferences, interaction patterns, and outcome 
feedback, the system can infer deeper user motivations and provide targeted 
interventions.

Third, we investigate human-like \textit{consolidation and forgetting} 
mechanisms. A dynamic memory lifecycle model will strengthen core memories and 
selectively forget outdated or redundant ones. Emotional memory modeling will 
enable more empathetic and humanized interaction.

\subsection{Summary}

In summary, the series of studies and practical deployments of the Memory Bear system—from laboratory performance validation to cross-industry applications in customer service, healthcare, and education—have fully demonstrated that introducing a human-like hierarchical and dynamic memory architecture is the key to overcoming the limitations of shallow interactions in artificial intelligence and achieving a qualitative leap in cognitive capability.

This memory framework not only enhances core cognitive dimensions such as contextual understanding, long-term decision-making, and cross-scenario adaptability, but also translates these technical advances into tangible industry benefits. In practical deployments, Memory Bear supports coherent interactions across more than 100 turns in intelligent customer service and enables the integration of historical symptoms with diagnostic reasoning in healthcare consultations. These capabilities move AI systems beyond simple response generation toward more precise and context-aware services, thereby significantly enhancing the practical utility and user trust.

More broadly, the technological breakthrough from ``passive memory’’ to ``active cognition’’ provides new foundational insights and a robust technical basis for the development of artificial general intelligence (AGI). A major bottlenecks of AGI lies in the absence of human-like memory management capabilities to support continuous learning and decision-making in complex environments. The ``multimodal encoding–intelligent maintenance–cognitive mapping’’  architecture of Memory Bear achieves a deep integration of memory and reasoning for the first time, offering a reusable engineering paradigm that breaks the long-standing reliance on parameter scaling in large model research.

With the open-sourcing of the Memory Bear core framework and the growing involvement of the developer community, we have strong reasons to believe that a new era of AI—one that truly understands user needs, accurately retains interaction memories, and continuously accompanies long-term user development—is rapidly approaching.

Over the next 3–5 years, memory-augmentation technologies represented by Memory Bear are highly likely to become the standard configuration for intelligent agent development. This shift will fundamentally transform the R\&D paradigm of intelligent systems—from ``pursuing parameter scale’’ to ``building efficient memory–cognition pipelines’’—ultimately steering artificial intelligence from ``high capability’’ toward ``high reliability, high adaptability, and high empathy.’’

\end{document}